\documentclass[11pt,a4paper]{article}
\usepackage[hyperref]{acl2018}
\usepackage{times}
\usepackage{latexsym}
\usepackage{url}
\usepackage{helvet}
\usepackage{courier}
\usepackage{amsmath}
\usepackage{graphicx}
\usepackage{fancyhdr}
\usepackage{adjustbox}
\usepackage{multirow}
\usepackage{float}
\usepackage{textcomp}
\usepackage{algorithm,algorithmic}
\usepackage{amssymb,amsthm}
\usepackage{bbm}

\aclfinalcopy 
\def\aclpaperid{1190} 

\title{Double Embeddings and CNN-based Sequence Labeling\\for Aspect Extraction}

\author{Hu Xu\textsuperscript{1}, Bing Liu\textsuperscript{1}, Lei Shu\textsuperscript{1}\and Philip S. Yu\textsuperscript{1,2}\\
\textsuperscript{1}{Department of Computer Science, University of Illinois at Chicago, Chicago, IL, USA}\\
\textsuperscript{2}{Institute for Data Science, Tsinghua University, Beijing, China}\\
{\tt \{hxu48, liub, lshu3, psyu\}@uic.edu}
}

\date{}

\begin{document}
\maketitle
\begin{abstract}
One key task of fine-grained sentiment analysis of product reviews is to extract product aspects or features that users have expressed opinions on. This paper focuses on supervised aspect extraction using deep learning. Unlike other highly sophisticated supervised deep learning models, this paper proposes a novel and yet simple CNN model \footnote{The code of this paper can be found at \url{https://www.cs.uic.edu/~hxu/}.} employing two types of pre-trained embeddings for aspect extraction: general-purpose embeddings and domain-specific embeddings. Without using any additional supervision, this model achieves surprisingly good results, outperforming state-of-the-art sophisticated existing methods. To our knowledge, this paper is the first to report such double embeddings based CNN model for aspect extraction and achieve very good results. 

\end{abstract}

\section{Introduction}
Aspect extraction is an important task in sentiment analysis \cite{HuL2004} and has many applications \cite{Liu2012}.
It aims to extract opinion targets (or aspects) from opinion text. 
In product reviews, aspects are product attributes or features. 
For example, from ``\textit{Its speed is incredible}'' in a laptop review, it aims to extract ``speed''. 

Aspect extraction has been performed using supervised \cite{Jakob2010,chernyshevich2014ihs,shu2017lifelong} and unsupervised approaches \cite{HuL2004, ZhuangJZ2006,MeiLWSZ2007,QiuLBC2011,yin2016unsupervised,he2017unsupervised}. 
Recently, supervised deep learning models achieved state-of-the-art performances \cite{li2017deep}. Many of these models use handcrafted features, lexicons, and complicated neural network architectures \cite{poria2016aspect,wang2016recursive,wang2017coupled,li2017deep}. 
Although these approaches can achieve better performances than their prior works, there are two other considerations that are also important.
(1) Automated feature (representation) learning is always preferred. 
How to achieve competitive performances without manually crafting features is an important question. 
(2) According to Occam's razor principle \cite{blumer1987occam}, a simple model is always preferred over a complex model.
This is especially important when the model is deployed in a real-life application (e.g., chatbot), where a complex model will slow down the speed of inference. Thus, to achieve competitive performance whereas keeping the model as simple as possible is important. This paper proposes such a model. 

To address the first consideration, we propose a double embeddings mechanism that is shown crucial for aspect extraction.
The embedding layer is the very first layer, where all the information about each word is encoded.
The quality of the embeddings determines how easily later layers (e.g., LSTM, CNN or attention) can decode useful information.
Existing deep learning models for aspect extraction use either a pre-trained general-purpose embedding, e.g., GloVe \cite{pennington2014glove}, or a general review embedding \cite{poria2016aspect}.
However, aspect extraction is a complex task that also requires fine-grained domain embeddings.
For example, in the previous example, detecting ``speed'' may require embeddings of both ``Its'' and ``speed''.
However, the criteria for good embeddings for ``Its'' and ``speed'' can be totally different.
``Its'' is a general word and the general embedding (trained from a large corpus) is likely to have a better representation for ``Its''.
But, ``speed'' has a very fine-grained meaning (e.g., how many instructions per second) in the \textit{laptop} domain, whereas ``speed'' in general embeddings or general review embeddings may mean how many miles per second.
So using in-domain embeddings is important even when the in-domain embedding corpus is not large. 
Thus, we leverage both general embeddings and domain embeddings and let the rest of the network to decide which embeddings have more useful information.

To address the second consideration, we use a pure Convolutional Neural Network (CNN) \cite{lecun1995convolutional} model for sequence labeling.
Although most existing models use LSTM \cite{hochreiter1997long} as the core building block to model sequences \cite{liu2015fine,li2017deep}, we noticed that CNN is also successful in many NLP tasks \cite{kim2014convolutional,zhang2015character,gehring2017convolutional}.
One major drawback of LSTM is that LSTM cells are sequentially dependent.
The forward pass and backpropagation must serially go through the whole sequence, which slows down the training/testing process
\footnote{We notice that a GPU with more cores has no training time gain on a low-dimensional LSTM because extra cores are idle and waiting for the other cores to sequentially compute cells.}.
One challenge of applying CNN on sequence labeling is that convolution and max-pooling operations are usually used for summarizing sequential inputs and the outputs are not well-aligned with the inputs. We discuss the solutions in Section \ref{sec:model}.

We call the proposed model \underline{D}ual \underline{E}mbeddings \underline{CNN} (DE-CNN).
To the best of our knowledge, this is the first paper that reports a double embedding mechanism and a pure CNN-based sequence labeling model for aspect extraction.

\section{Related Work}
Sentiment analysis has been studied at document, sentence and aspect levels \cite{Liu2012,Pang2008OMS,Cambria2012}. This work focuses on the aspect level \cite{HuL2004}. Aspect extraction is one of its key tasks, and has been performed using both unsupervised and supervised approaches. 
The unsupervised approach includes methods such as frequent pattern mining \cite{HuL2004,PopescuNE2005}, syntactic rules-based extraction \cite{ZhuangJZ2006,WangBo2008,QiuLBC2011}, topic modeling \cite{MeiLWSZ2007,TitovM2008,Lin2009,Moghaddam2011}, word alignment \cite{KangLiu2013IJCAI} and label propagation \cite{Zhou-wan-xiao:2013:EMNLP}.

Traditionally, the supervised approach \cite{Jakob2010,Mitchell-EtAl:2013:EMNLP,shu2017lifelong} uses Conditional Random Fields (CRF) \cite{Lafferty2001conditional}.
Recently, deep neural networks are applied to learn better features for supervised aspect extraction, e.g., using
LSTM \cite{williams1989learning,hochreiter1997long,liu2015fine} and
attention mechanism \cite{wang2017coupled,he2017unsupervised} together with manual features \cite{poria2016aspect,wang2016recursive}.
Further, \cite{wang2016recursive,wang2017coupled,li2017deep} also proposed aspect and opinion terms co-extraction via a deep network.
They took advantage of the gold-standard opinion terms or sentiment lexicon for aspect extraction.
The proposed approach is close to \cite{liu2015fine}, where only the annotated data for aspect extraction is used. 
However, we will show that our approach is more effective even compared with baselines using additional supervisions and/or resources.

The proposed embedding mechanism is related to cross domain embeddings \cite{bollegala2015unsupervised,bollegala2017think} and domain-specific embeddings \cite{xumeta,Xu2018pro}. 
However, we require the domain of the domain embeddings must exactly match the domain of the aspect extraction task. 
CNN \cite{lecun1995convolutional,kim2014convolutional} is recently adopted for named entity recognition \cite{strubell2017fast}.
CNN classifiers are also used in sentiment analysis \cite{poria2016aspect,chen2017improving}.
We adopt CNN for sequence labeling for aspect extraction because CNN is simple and parallelized.

\section{Model}
\label{sec:model}

\begin{figure}[t]
\centering    
\includegraphics[width=3in]{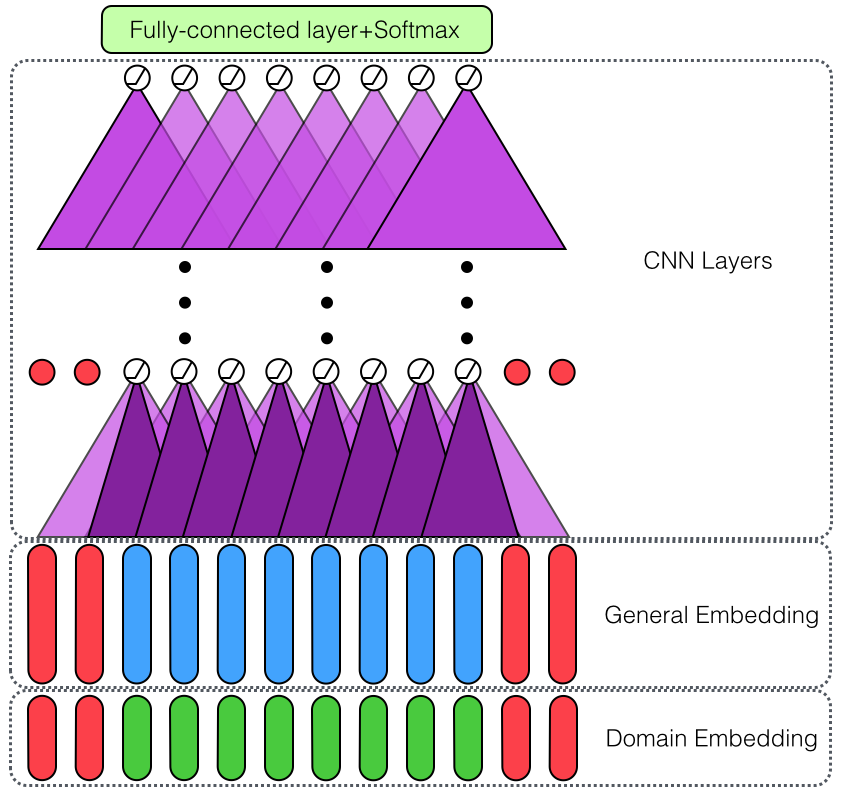}
    \caption{Overview of DE-CNN: red vectors are zero vectors; purple triangles are CNN filters. }
    \label{fig:fr}
\end{figure}

The proposed model is depicted in Figure \ref{fig:fr}.
It has 2 embedding layers, 4 CNN layers, a fully-connected layer shared across all positions of words, and a softmax layer over the labeling space $\mathcal{Y}=\{B, I, O\}$ for each position of inputs.
Note that an aspect can be a phrase and $B$, $I$ indicate the beginning word and non-beginning word of an aspect phrase and $O$ indicates non-aspect words.

Assume the input is a sequence of word indexes $\mathbf{x}=(x_1, \dots, x_n)$.
This sequence gets its two corresponding continuous representations $\mathbf{x}^g$ and $\mathbf{x}^d$ via two separate embedding layers (or embedding matrices) $W^g$ and $W^d$.
The first embedding matrix $W^g$ represents general embeddings pre-trained from a very large general-purpose corpus (usually hundreds of billions of tokens).
The second embedding matrix $W^d$ represents domain embeddings pre-trained from a small in-domain corpus, where the scope of the domain is exactly the domain that the training/testing data belongs to.
As a counter-example, if the training/testing data is in the \textit{laptop} domain, then embeddings from the \textit{electronics} domain are considered to be out-of-domain embeddings (e.g., the word ``adapter'' may represent different types of adapters in \textit{electronics} rather than exactly a \textit{laptop} adapter). That is, only laptop reviews are considered to be in-domain. 

We do not allow these two embedding layers trainable because small training examples may lead to many unseen words in test data.
If embeddings are tunable, the features for seen words' embeddings will be adjusted (e.g., forgetting useless features and infusing new features that are related to the labels of the training examples).
And the CNN filters will adjust to the new features accordingly. 
But the embeddings of unseen words from test data still have the old features that may be mistakenly extracted by CNN.

Then we concatenate two embeddings $\mathbf{x}^{(1)}=\mathbf{x}^g \oplus \mathbf{x}^d$ and feed the result into a stack of 4 CNN layers. A CNN layer has many 1D-convolution filters and each (the $r$-th) filter has a fixed kernel size $k=2c+1$ and performs the following convolution operation and ReLU activation: 
\begin{equation}
x_{i,r}^{(l+1)}=\max\bigg(0, (\sum_{j=-c}^c w_{j,r}^{(l)} x_{i+j}^{(l)})+b_r^{(l)}\bigg),
\end{equation}
where $l$ indicates the $l$-th CNN layer. 
We apply each filter to all positions $i=1:n$.
So each filter computes the representation for the $i$-th word along with $2c$ nearby words in its context.  
Note that we force the kernel size $k$ to be an odd number and set the stride step to be 1 and further pad the left $c$ and right $c$ positions with all zeros.  
In this way, the output of each layer is well-aligned with the original input $\mathbf{x}$ for sequence labeling purposes.
For the first ($l=1$) CNN layer, we employ two different filter sizes. 
For the rest 3 CNN ($l \in \{2, 3, 4\}$) layers, we only use one filter size.
We will discuss the details of the hyper-parameters in the experiment section.
Finally, we apply a fully-connected layer with weights shared across all positions and a softmax layer to compute label distribution for each word.
The output size of the fully-connected layer is $|\mathcal{Y}|=3$.
We apply dropout after the embedding layer and each ReLU activation.
Note that we do not apply any max-pooling layer after convolution layers because a sequence labeling model needs good representations for every position and max-pooling operation mixes the representations of different positions, which is undesirable (we show a max-pooling baseline in the next section).

\section{Experiments}

\subsection{Datasets}
\begin{table}[t]
    \label{tab:dataset} 
    \centering
    \scalebox{0.8}{
        \begin{tabular}{c|c|c}
        \hline
            {\bf Description}  &{\bf Training }        &{\bf Testing }  \\
                               &{\bf \#S./\#A.} &{\bf \#S./\#A.}  \\\hline
            SemEval-14 Laptop  &3045/2358              &800/654\\\hline
            SemEval-16 Restaurant&2000/1743            &676/622\\\hline
        \end{tabular}
    }
    \caption{Dataset description with the number of sentences(\#S.) and number of aspect terms(\#A.)}
\end{table}

Following the experiments of a recent aspect extraction paper \cite{li2017deep},
we conduct experiments on two benchmark datasets from SemEval challenges \cite{pontiki2014SemEval,pontiki2016semeval} as shown in Table \ref{tab:dataset}. 
The first dataset is from the \textit{laptop} domain on subtask 1 of SemEval-2014 Task 4.
The second dataset is from the \textit{restaurant} domain on subtask 1 (slot 2) of SemEval-2016 Task 5.
These two datasets consist of review sentences with aspect terms labeled as spans of characters.
We use NLTK\footnote{\url{http://www.nltk.org/} } to tokenize each sentence into a sequence of words. 

For the general-purpose embeddings, we use the glove.840B.300d embeddings \cite{pennington2014glove}, which are pre-trained from a corpus of 840 billion tokens that cover almost all web pages. These embeddings have 300 dimensions.
For domain-specific embeddings, we collect a laptop review corpus and a restaurant review corpus and use fastText \cite{bojanowski2016enriching} to train domain embeddings.  
The laptop review corpus contains all laptop reviews from the Amazon Review Dataset \cite{he2016ups}.
The restaurant review corpus is from the Yelp Review Dataset Challenge \footnote{\url{https://www.yelp.com/dataset/challenge} }.
We only use reviews from restaurant categories that the second dataset is selected from \footnote{\url{http://www.cs.cmu.edu/~mehrbod/RR/Cuisines.wht} }.
We set the embedding dimensions to 100 and the number of iterations to 30 (for a small embedding corpus, embeddings tend to be under-fitted), and keep the rest hyper-parameters as the defaults in fastText.
We further use fastText to compose out-of-vocabulary word embeddings via subword N-gram embeddings.

\subsection{Baseline Methods}
We perform a comparison of DE-CNN with three groups of baselines using the standard evaluation of the datasets\footnote{\url{http://alt.qcri.org/semeval2014/task4}} \footnote{\url{http://alt.qcri.org/semeval2016/task5}}.
The results of the first two groups are copied from \cite{li2017deep}.
The first group uses single-task approaches.

\textbf{CRF} is conditional random fields with basic features\footnote{\url{http://sklearn-crfsuite.readthedocs.io/en/latest/tutorial.html} } and GloVe word embedding\cite{pennington2014glove}.

\textbf{IHS\_RD} \cite{chernyshevich2014ihs} and \textbf{NLANGP} \cite{toh2016nlangp} are best systems in the original challenges \cite{pontiki2014SemEval,pontiki2016semeval}.

\textbf{WDEmb} \cite{yin2016unsupervised} enhanced CRF with word embeddings, linear context embeddings and dependency path embeddings as input.

\textbf{LSTM} \cite{liu2015fine,li2017deep} is a vanilla BiLSTM.

\textbf{BiLSTM-CNN-CRF} \cite{Reimers:2017:EMNLP} is the state-of-the-art from the Named Entity Recogntion (NER) community. We use this baseline\footnote{\url{https://github.com/UKPLab/emnlp2017-bilstm-cnn-crf} } to demonstrate that a NER model may need further adaptation for aspect extraction.

The second group uses multi-task learning and also take advantage of gold-standard opinion terms/sentiment lexicon.

\textbf{RNCRF} \cite{wang2016recursive} is a joint model with a dependency tree based recursive neural network and CRF for aspect and opinion terms co-extraction. 
Besides opinion annotations, it also uses handcrafted features.

\textbf{CMLA} \cite{wang2017coupled} is a multi-layer coupled-attention network that also performs aspect and opinion terms co-extraction. It uses gold-standard opinion labels in the training data.

\textbf{MIN} \cite{li2017deep} is a multi-task learning framework that has (1) two LSTMs for jointly extraction of aspects and opinions, and (2) a third LSTM for discriminating sentimental and non-sentimental sentences. 
A sentiment lexicon and high precision dependency rules are employed to find opinion terms. 

The third group is the variations of DE-CNN.

\textbf{GloVe-CNN} only uses glove.840B.300d to show that domain embeddings are important. 

\textbf{Domain-CNN} does not use the general embeddings to show that domain embeddings alone are not good enough as the domain corpus is limited for training good general words embeddings.

\textbf{MaxPool-DE-CNN} adds max-pooling in the last CNN layer. We use this baseline to show that the max-pooling operation used in the traditional CNN architecture is harmful to sequence labeling.

\textbf{DE-OOD-CNN} replaces the domain embeddings with out-of-domain embeddings to show that a large out-of-domain corpus is not a good replacement for a small in-domain corpus for domain embeddings.
We use all \textit{electronics} reviews as the out-of-domain corpus for the \textit{laptop} and all the Yelp reviews for \textit{restaurant}.

\textbf{DE-Google-CNN} replaces the glove embeddings with GoogleNews embeddings\footnote{\url{https://code.google.com/archive/p/word2vec/} }, which are pre-trained from a smaller corpus (100 billion tokens). We use this baseline to demonstrate that general embeddings that are pre-trained from a larger corpus performs better.

\textbf{DE-CNN-CRF} replaces the softmax activation with a CRF layer\footnote{\url{https://github.com/allenai/allennlp}}. We use this baseline to demonstrate that CRF may not further improve the challenging performance of aspect extraction.

\subsection{Hyper-parameters}
We hold out 150 training examples as validation data to decide the hyper-parameters.
The first CNN layer has 128 filters with kernel sizes $k=3$ (where $c=1$ is the number of words on the left (or right) context) and 128 filters with kernel sizes $k=5$ ($c=2$).
The rest 3 CNN layers have 256 filters with kernel sizes $k=5$ ($c=2$) per layer.
The dropout rate is 0.55 and the learning rate of Adam optimizer \cite{kingma2014adam} is 0.0001 because CNN training tends to be unstable.

\begin{table}[t]
    \label{tab:result} 
    \centering
    \scalebox{0.85}{
        \begin{tabular}{c||c|c}
        \hline
        {\bf Model} &{\bf Laptop }  &{\bf Restaurant }  \\\hline
        CRF         &74.01      &69.56  \\
        IHS\_RD     &74.55      &-      \\
        NLANGP      &-          &72.34  \\
        WDEmb       &75.16      &-      \\
        LSTM        &75.25      &71.26  \\
		BiLSTM-CNN-CRF &77.8 & 72.5\\
        \hline
        RNCRF       &78.42 &-      \\
        CMLA        &77.80      &-      \\
        MIN         &77.58      &73.44  \\
        \hline
        \hline
        GloVe-CNN & 77.67 & 72.08\\
        Domain-CNN & 78.12 & 71.75\\
        MaxPool-DE-CNN & 77.45 & 71.12\\
        DE-LSTM & 78.73 & 72.94 \\
        DE-OOD-CNN & 80.21 & 74.2 \\
		DE-Google-CNN & 78.8 & 72.1 \\
		DE-CNN-CRF & 80.8 & 74.1 \\
        DE-CNN &\textbf{81.59}* &\textbf{74.37}* \\
        \hline
        \end{tabular}
    }
    \caption{Comparison results in F$_1$ score: numbers in the third group are averaged scores of 5 runs. * indicates the result is statistical significant at the level of 0.05.}
\end{table}

\subsection{Results and Analysis}
Table \ref{tab:result} shows that DE-CNN performs the best. 
The double embedding mechanism improves the performance and in-domain embeddings are important. 
We can see that using general embeddings (GloVe-CNN) or domain embeddings (Domain-CNN) alone gives inferior performance. 
We further notice that the performance on \textit{Laptops} and \textit{Restaurant} domains are quite different. 
\textit{Laptops} has many domain-specific aspects, such as ``adapter''. 
So the domain embeddings for \textit{Laptops} are better than the general embeddings. 
The \textit{Restaurant} domain has many very general aspects like ``staff'', ``service'' that do not deviate much from their general meanings. 
So general embeddings are not bad. 
Max pooling is a bad operation as indicated by MaxPool-DE-CNN since the max pooling operation loses word positions.
DE-OOD-CNN's performance is poor, indicating that making the training corpus of domain embeddings to be exactly in-domain is important.
DE-Google-CNN uses a much smaller training corpus for general embeddings, leading to poorer performance than that of DE-CNN.
Surprisingly, we notice that the CRF layer (DE-CNN-CRF) does not help.
In fact, the CRF layer can improve 1-2\% when the laptop's performance is about 75\%.
But it doesn't contribute much when laptop's performance is above 80\%. 
CRF is good at modeling label dependences (e.g., label $I$ must be after $B$), but many aspects are just single words and the major types of errors (mentioned later) do not fall in what CRF can solve.
Note that we did not tune the hyperparameters of DE-CNN-CRF for practical purpose because training the CRF layer is extremely slow. 

One important baseline is BiLSTM-CNN-CRF, which is markedly worse than our method. 
We believe the reason is that this baseline leverages dependency-based embeddings\cite{levy2014dependency}, 
which could be very important for NER.
NER models may require further adaptations (e.g., domain embeddings) for opinion texts. 

DE-CNN has two major types of errors.
One type comes from inconsistent labeling (e.g., for the restaurant data, the same aspect is sometimes labeled and sometimes not). 
Another major type of errors comes from unseen aspects in test data that require the semantics of the conjunction word ``and'' to extract. For example, if A is an aspect and when ``A and B'' appears, B should also be extracted but not.
We leave this to future work.

\section{Conclusion}
We propose a CNN-based aspect extraction model with a double embeddings mechanism without extra supervision.
Experimental results demonstrated that the proposed method outperforms state-of-the-art methods with a large margin.

\section{ Acknowledgments}
This work was supported in part by NSF through grants IIS-1526499, IIS-1763325, and IIS1407927, CNS-1626432, NSFC 61672313, and a gift from Huawei Technologies.

\bibliography{acl2018}
\bibliographystyle{acl_natbib}

\end{document}